\documentclass{esannV2}
\usepackage[dvips]{graphicx}
\usepackage[latin1]{inputenc}
\usepackage{amssymb,amsmath,array}
\usepackage{booktabs}
\usepackage{url}
\usepackage{adjustbox}
\usepackage{multirow}
\usepackage{subcaption}
\usepackage{float}

\usepackage{ltxcmds}

\usepackage{xspace}
\usepackage{hyperref}
\hypersetup{
  breaklinks=true,   % splits links across lines
  colorlinks=true,   % displays links as colored text instead of blocks
  pdfusetitle=true,  % \title and \author values into pdf metadata
  linkcolor={blue},
  urlcolor={black},
  citecolor={black}
}

\newcommand{\ie}{i.\,e.,\xspace}

\usepackage{bm}
\def\eqref#1{equation~\ref{#1}}
\def\1{\bm{1}}

\def\vd{{\bm{d}}}

\def\mH{{\bm{H}}}
\def\mI{{\bm{I}}}

\def\mN{{\bm{N}}}

\def\mR{{\bm{R}}}

\def\mU{{\bm{U}}}

\def\mW{{\bm{W}}}

\def\mZ{{\bm{Z}}}

\def\mLambda{{\bm{\Lambda}}}
\def\mSigma{{\bm{\Sigma}}}

\DeclareMathAlphabet{\mathsfit}{\encodingdefault}{\sfdefault}{m}{sl}
\SetMathAlphabet{\mathsfit}{bold}{\encodingdefault}{\sfdefault}{bx}{n}

\begin{document}
%style file for ESANN manuscripts

\title{Isotropy Matters: Soft-ZCA Whitening of Embeddings for Semantic Code Search}
%\title{Improving Semantic Code Search via Isotropy-Aware Embedding Space Whitening}
%\title{Representation Space of Code Language Models / Finding the Optimal Isotropy for Semantic Code Search}

%***********************************************************************
% AUTHORS INFORMATION AREA
%***********************************************************************
\author{Andor Diera$^1$ and Lukas Galke$^2$ and Ansgar Scherp$^1$
%
% Optional short acknowledgment: remove next line if non-needed
\thanks{This work was co-funded by DFG as part of the CodeInspector
Project - P5311038}
%
% DO NOT MODIFY THE FOLLOWING '\vspace' ARGUMENT
\vspace{.3cm}\\
%
% Addresses and institutions (remove "1- " in case of a single institution)
1- Ulm University - Data Science and Big Data Analytics \\
Helmholtzstrasse 16, 89081 Ulm - Germany
%
% Remove the next three lines in case of a single institution
\vspace{.1cm}\\
2- University of Southern Denmark - Dept of Mathematics and Computer Science \\
Campusvej 55, 5230 Odense - Denmark\\
}
%***********************************************************************
% END OF AUTHORS INFORMATION AREA
%***********************************************************************

\maketitle

\begin{abstract}
Low isotropy in an embedding space impairs performance on tasks involving semantic inference. Our study investigates the impact of isotropy on semantic code search performance and explores post-processing techniques to mitigate this issue. We analyze various code language models, examine isotropy in their embedding spaces, and its influence on search effectiveness. We propose a modified ZCA whitening technique to control isotropy levels in embeddings. Our results demonstrate that Soft-ZCA whitening improves the performance of pre-trained code language models and can complement contrastive fine-tuning.
\end{abstract}

\section{Introduction}

Isotropy in language models (LMs) refers to the uniform distribution of vector representations in the embedding space~\cite{cai2020isotropy}. 
It enhances the efficient use of the embedding space and increases robustness to perturbations. % ~\cite{cai2020isotropy}.
Anisotropy, \ie when vectors are unevenly distributed, can hinder model performance on semantic tasks by making it difficult to distinguish between different meanings~\cite{attieh2023optimizing}. 
An anisotropic embedding space poses an even greater challenge for cross-lingual tasks, where accurate semantic alignment demands more precise representational distinctions~\cite{ji2023isotropic}.
These representational challenges extend to code LMs, notably affecting semantic code search, where natural language queries are used to retrieve relevant code snippets~\cite{husain2019codesearchnet}. In this task, high anisotropy can lead to suboptimal retrieval performance, as the encoded representations of semantically different code snippets may not be adequately distinguished. The prevalence of programming language keywords and symbols intensifies this problem, as these elements can dominate the sequence representations and obscure the semantic content of the code~\cite{eghbali2022crystalbleu}. %Addressing this issue is important for improving the effectiveness of semantic code search, where precise and meaningful representation of code semantics is essential for achieving optimal search results.

Contrastive fine-tuning is a common approach for improving semantic code search by encouraging the model to bring semantically similar code representations closer while pushing dissimilar ones apart~\cite{wang2023codet5+, diera2023gencodesearchnet}.
However, fine-tuning only marginally mitigates the anisotropy problem, as it does not fully address the underlying issue of the generally low angular distance between encoded representations~\cite{rajaee-pilehvar-2021-fine-tuning}. 
In NLP, multiple approaches have been proposed to improve the isotropy of an embedding space. Regularization methods~\cite{ji2023isotropic,attieh2023optimizing} and simple post-processing techniques~\cite{cai2020isotropy, su2021whitening} have shown promise in enhancing the isotropy of encoded representations. ZCA whitening~\cite{bell1996edges} has shown to be a particularly fitting post-processing method for decorrelating the hidden features and increasing the isotropy of embeddings~\cite{su2021whitening}.
As vector databases become increasingly central to modern search systems, there is growing interest in lightweight post-processing techniques that can boost performance. 
However, these techniques are unexplored in the context of code search tasks.
To address the challenge of anisotropy in semantic code search, we analyze the embedding space of three pre-trained code LMs: CodeBERT~\cite{feng2020codebert}, CodeT5+~\cite{wang2023codet5+}, and Code Llama~\cite{roziere2023code}. We examine how evenly distributed (isotropic) are their embeddings, specifically looking at how this affects their performance on code search. 
We introduce Soft-ZCA, an extension to ZCA whitening which permits control over the degree of whitening. 
We evaluate our approach on six popular programming languages and test the generalization capabilities on a low-resource programming language dataset. Our analysis shows that, similarly to standard LMs, code LMs also showcase a high level of anisotropy. We confirm that contrastive fine-tuning does not have a strong effect on isotropy and demonstrate that applying Soft-ZCA whitening with an eigenvalue regularizer can improve both isotropy and code search performance. The code for our experiments is available at \url{https://github.com/drndr/code\_isotropy}.
The main contributions of our paper are:
\begin{itemize}
    \item We analyze the isotropy of the embedding spaces in three popular code LMs regarding code search performance.
    \item We introduce a regularizer to ZCA whitening to control the degree of isotropy in embeddings, which we call Soft-ZCA.
    \item Experiments on six popular and one low-resource programming language show that post-processing the embeddings with Soft-ZCA whitening improves code search for pre-trained and fine-tuned code LMs. 
\end{itemize}

\section{Whitening of the Embeddings}
\label{sec:methods}

%\todoyellow{we need to harmonize the use of notation, sometimes matrices are $Z$ and in other cases, it is $\mathbf{Z}$. a: my current system is $Z$ is for definition, $\mathbf{Z}$ is for equation - for better readability, we can discuss this.

%The mathcommans.tex have predefined formatting for that, \ie use the command from the Deep Learning book to print $\mZ$ (m in the command is for matrix)}

Whitening is a common processing step in machine learning and statistical analysis to transform variables or features to orthogonality~\cite{kessy2018optimal}. 
Given a set of embeddings $\mZ \in \mathbb{\mR}^{\mN \times \vd}$, a whitening transformation can be denoted as $\mH = \mW \mZ^\top$, where $\mW \in \mathbb{\mR}^{\vd \times \vd}$ is the square whitening matrix, and $\mH \in \mathbb{R}^{\mN \times \vd}$ is the whitened embedding. Since the only condition of a whitening transformation is to satisfy $\mW \mSigma \mW^\top = \mI$ (where $\mSigma$ is the covariance matrix of $\mZ$), there are infinitely many possible whitening transformations due to rotational freedom. 
In practice, the most widely used whitening transformations are based on Principal Component Analysis (PCA), Zero-phase Component Analysis (ZCA), or the Cholesky decomposition of the covariance matrix, each offering different properties and tradeoffs for various applications~\cite{kessy2018optimal}.
ZCA whitening~\cite{bell1996edges} has been shown to maintain the highest correlation with the original data~\cite{kessy2018optimal} and is considered the most appropriate for embedding spaces. 
The whitening matrix of ZCA is defined as $\mW^{\operatorname{ZCA}} = \mSigma^{-1/2}$. Using singular value decomposition, $\mSigma^{-1/2}$ can be rewritten to $\mSigma^{-1/2} = \mU \mLambda^{-1/2} \mU^\top$
where $\mU$ is an orthogonal matrix based on the eigenvectors of $\mSigma$, and $\mLambda$ is a diagonal matrix of $\mSigma$'s eigenvalues. 

\paragraph{Soft-ZCA Whitening}To control the degree of whitening we introduce an eigenvalue regularizer $\epsilon$. This adjustment modifies the whitening matrix calculation to $\mW^{\operatorname{ZCA}} = \mU( \mLambda + \epsilon\mI)^{-1/2} \mU^\top$, where $\mI$ is the identity matrix. The key purpose of $\epsilon$ is to retain more of the original signal and variance. If any of the eigenvalues in $\mLambda$ are close to 0, their inverse square root will become exceedingly large, which causes the whitening transformation to amplify noise and insignificant components in the data. By placing a lower bound on the eigenvalues of $\mSigma^{-1/2}$, $\epsilon$ can directly influence the strength of the whitening transformation.

\section{Experimental Apparatus}
\label{sec:experimentalapparatus}

\paragraph{Datasets}
\label{sec:datasets}
We experiment using two datasets.
The first is CodeSearchNet~\cite{husain2019codesearchnet}, a benchmark for studying the code search capabilities of machine learning models. It encompasses code-comment pairs from six popular programming languages: Python, Go, Java, JavaScript, Ruby, and PHP. The full corpus comprises 2 million code-comment pairs. To evaluate generalization to a low-resource language, we use the StatCodeSearch test dataset~\cite{diera2023gencodesearchnet}, which comprises
1,070 code-comment pairs from social science research in the R language.

\paragraph{Models}
We investigate the embeddings of three code LMs. CodeBERT~\cite{feng2020codebert} is an encoder-only LM developed for programming language understanding. CodeT5+~\cite{wang2023codet5+} is an encoder-decoder LM trained for both code understanding and generation. Code Llama~\cite{roziere2023code} is the code-specialized version of Llama 2, used predominantly for generation.
Apart from the base models (pre-trained only), we also include a fine-tuned CodeBERT that is trained on the code-comment pairs for each search task.
%
%The important 
Characteristics of each model can be seen in Table~\ref{tab:model_details}.

\begin{table}[ht!]
    %\small
    \caption{Model details}
    \centering  
    \begin{adjustbox}{height=1.2cm}
    \begin{tabular}{l|rrrr}
        \toprule
        %Model & \#Parameters & Embedding dim. & \#Progr. lang. & Contrastive Pre-train\\
          Model & \vtop{\hbox{\strut Number of }\hbox{\strut Parameters}}   & \vtop{\hbox{\strut Embedding }\hbox{\strut Dimension}}  & \vtop{\hbox{\strut Supported }\hbox{\strut Progr. lang.}} & \vtop{\hbox{\strut Contrastive }\hbox{\strut Pre-training}}\\
         \midrule
         CodeBERT & 125m & 768 & 6 & no \\
         CodeT5+ & 110m & 256 & 9 & yes\\
         Code Llama & 7b & 4,096 & 7 & no \\
         \bottomrule
    \end{tabular}
    \end{adjustbox}
    \label{tab:model_details}
\end{table}

%\subsection{Measures}
%\label{sec:measures}

%\paragraph{Isotropy}
%To measure the isotropy of the embedding space, we employ the IsoScore metric~\cite{rudman2022isoscore}. Compared to other isotropy measures (i.e. average cosine similarity), this metric relies on the mathematical definition of isotropy, and fulfills the essential properties for accurate measurement. Namely, it is mean agnostic, rotational invariant, invariant to scalar changes to the covariance matrix but sensitive to the maximum variance, and scales linearly with the number of dimensions used. IsoScore values are bounded to [0,1], where 1 indicates perfect isotropy, while 0 represents a perfectly anisotropic space.

%\paragraph{Code Search}
%To evaluate the ranking in the code search task, we the employ Mean Reciprocal Rank~(MRR) metric. For each query (e.g. code comment) we measure the cosine distance to all code snippets in the dataset and create a ranked list based on it. The reciprocal rank of the correct code snippet is used for each query to compute the metric. Formally, for a set of queries~$Q$, the reciprocal of the best-ranked correct answer at rank~$r_i$ is aggregated and averaged as $\mathrm{MRR} = \frac{1}{|Q|} \sum_{i=1,\ldots, |Q|} \frac{1}{r_i}\,$.

\paragraph{Procedure}
\label{sec:procedure}
We process code and natural language inputs independently, mirroring real-world search systems. Each sequence is cut off at 256 tokens with no padding added. For CodeBERT and Code LLama, we extract the sequence representations by applying mean pooling on the last hidden state.
For the CodeT5+ model, we rely on the default pooling, which includes an additional down-projection. 
We fine-tune CodeBERT for each dataset separately with InfoNCE as a contrastive loss, a learning rate of 5e-5, and a batch size of 32 for 5 epochs. The whitening matrices are calculated independently for code and comments, using the full test sets. We employ IsoScore~\cite{rudman2022isoscore} to measure the isotropy of the embedding space. It is bounded to $[0,1]$, where $1$ indicates perfect isotropy. %, while 0 represents a anisotropic space. 
Code search is evaluated using the Mean Reciprocal Rank (MRR) based on the cosine distance between the comments and codes. For each programming language, we rank all codes for each comment in the test set.

\section{Results}
\label{sec:results}

To evaluate the embedding space of the models, we first measure their isotropy and ranking performance. To better assess the difference between natural and programming language, isotropy is measured separately for code and comment representations. Table \ref{tab:baseline} presents the MRR and IsoScores.

In summary, CodeT5+ achieves the highest MRR and IsoScores, surpassing even fine-tuned (FT) CodeBERT models. While fine-tuning greatly improves CodeBERT's ranking performance, its impact on isotropy is minor, with an average IsoScore increase of 0.073. Supplementary experiments with disabling the down-projection in CodeT5+ show worse results than fine-tuning CodeBERT, demonstrating that using a smaller hidden dimension in itself benefits both isotropy and ranking performance. Overall, our analysis shows that while models that have higher isotropy perform better, the relationship between MRR and IsoScore is not linear. Additionally, the analysis reveals that the isotropy of code and comment embeddings differs only marginally. This suggests that code and comment embeddings can be treated as similarly isotropic in practice.

\begin{figure*}[t!]
    \centering
    \begin{subfigure}[t]{0.45\textwidth} % Use fixed width 4cm here
        \centering
        \includegraphics[height=1.65in]{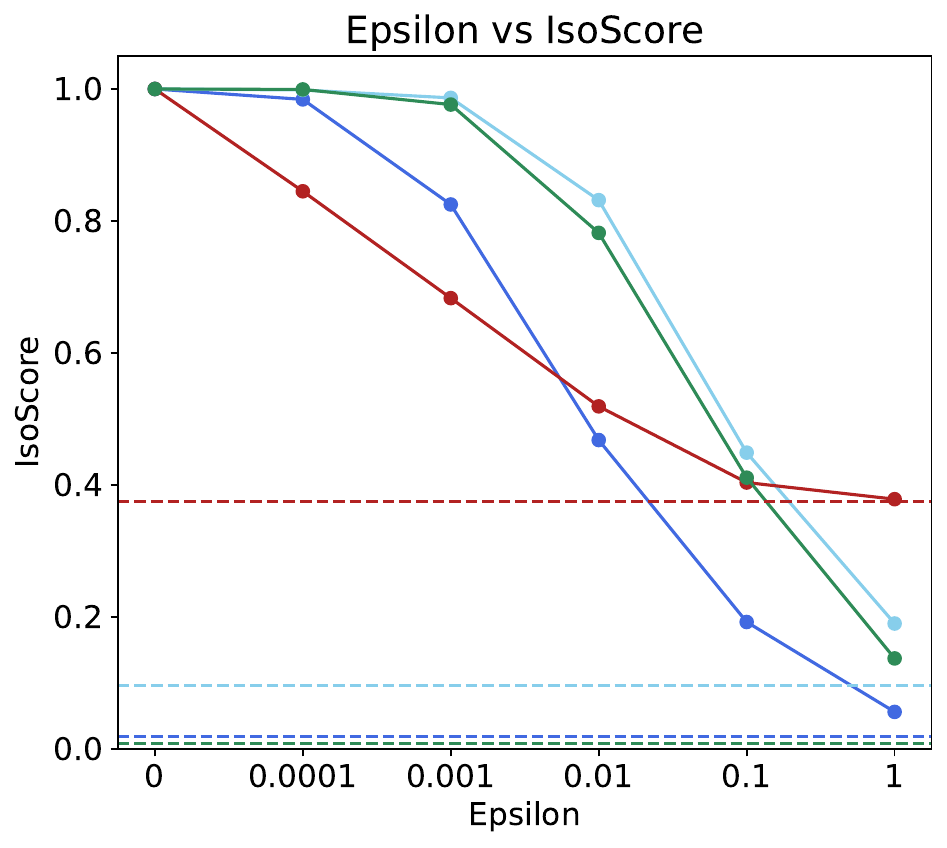}
        %\caption{Isoscore}
    \end{subfigure}%
    \begin{subfigure}[t]{0.54\textwidth}
        \centering
        \includegraphics[height=1.65in]{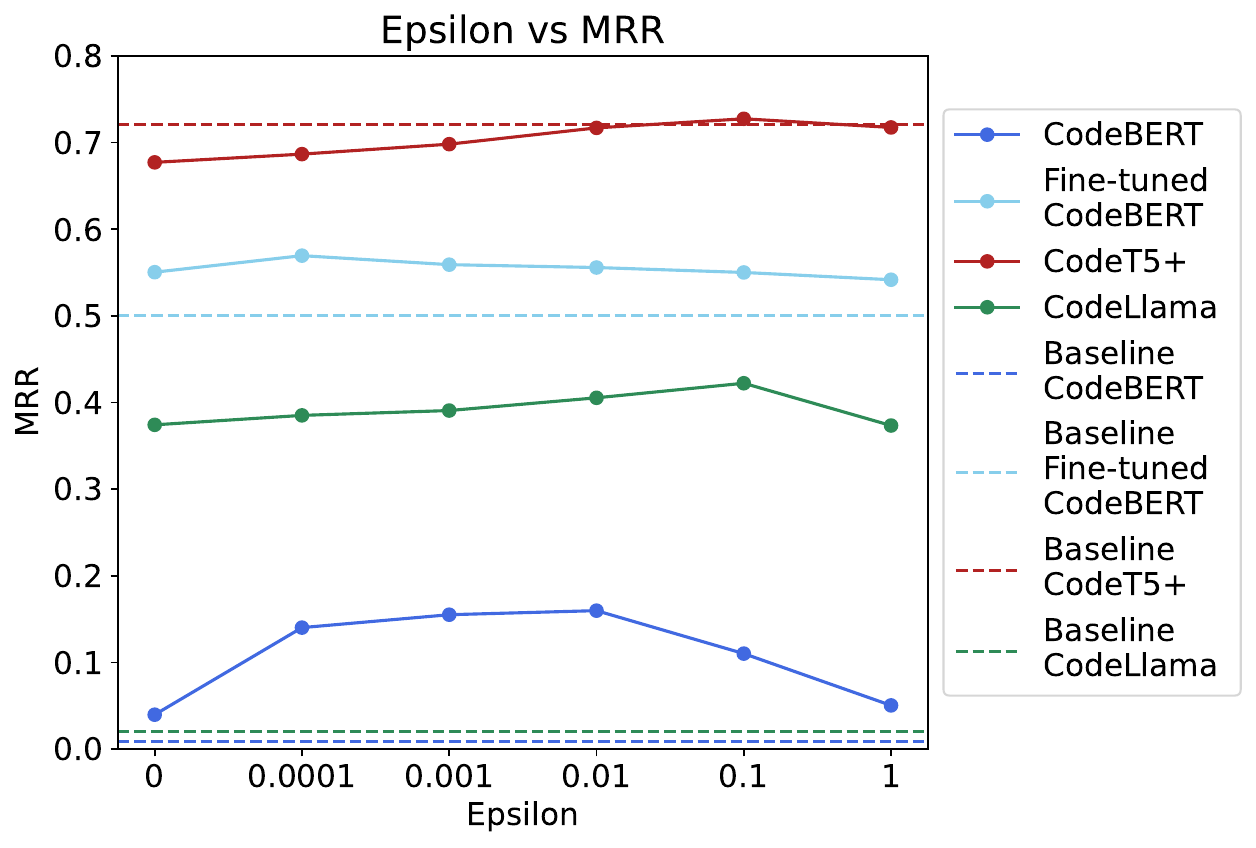}
        %\caption{M}
    \end{subfigure}
    \caption{Average IsoScore (left) and MRR measures (right) at different epsilon values on the CodeSearchNet Python dataset}
    \label{fig:epsilon}
\end{figure*}

Applying standard ZCA whitening (where $\epsilon=0$) greatly improves the base CodeBERT and Code Llama results, but in the case of fine-tuned CodeBERT and CodeT5+ it decreased the ranking performance on most datasets. With the introduction of the eigenvalue regularizer, we found that moderate whitening ($\epsilon\in \{0.1,0.01\}$) results in the best performance with the base models and only the fine-tuned CodeBERT requires stronger whitening($\epsilon=0.0001$) for optimal performance. Figure~\ref{fig:epsilon} showcases the interaction between the eigenvalue regularizer and Isoscore/MRR. 
Overall, we find that the optimal IsoScore for the base models ranges between $0.2$ and $0.8$, while the fine-tuned model performs best with almost perfect isotropy. This pattern is demonstrated in Table~\ref{tab:gains}, where fine-tuned CodeBERT achieves IsoScores consistently above $0.99$ across all programming languages while delivering moderate MRR improvements (ranging from$ +0.042$ to $+0.075$), whereas Code Llama shows more substantial MRR gains (up to $+0.476$ for Ruby) with IsoScores between $0.224$ and $0.496$. 
Importantly, the positive $\Delta$ MRR values across nearly all models and programming languages demonstrate that the Soft-ZCA whitening technique can reliably improve code search performance. This suggests the technique is robust and effective across different model architectures and provides performance benefits even for low-resource programming languages not present in the training data.

\begin{table}[h!]
    %\small
    \centering
    \caption{MRR and IsoScores (Code\,/\,Comment) on the CodeSearchNet and StatCodeSearch(R) datasets using non-whitened embeddings.}
    \begin{adjustbox}{width=1\textwidth}
    \begin{tabular}{l|cc|cc|cc|cc}
        \toprule
         & 
         \multicolumn{2}{c}{CodeBERT} &
         \multicolumn{2}{c}{FT CodeBERT} &
         \multicolumn{2}{c}{CodeT5+} &
         \multicolumn{2}{c}{Code LLama} \\
         \midrule
         & MRR & IsoScores & MRR & IsoScores & MRR & IsoScores & MRR & IsoScores \\
         \midrule
         Ruby & 0.006 & 0.007\,/\,0.014 & 0.547 & 0.052\,/\,0.062 & 0.705 & 0.350\,/\,0.296 & 0.047 & 0.008\,/\,0.003 \\
         Javascript  & 0.002 & 0.005\,/\,0.013 & 0.427 & 0.065\,/\,0.072 &  0.638 & 0.365\,/\,0.335 & 0.026 & 0.010\,/\,0.002 \\
         Go & 0.002 & 0.006\,/\,0.010 & 0.619 & 0.050\,/\,0.036&  0.757 & 0.234\,/\,0.196& 0.031 & 0.006\,/\,0.003\\
         Java  & 0.000 & 0.007\,/\,0.007 & 0.395 & 0.059\,/\,0.067 & 0.595 & 0.388\,/\,0.313 & 0.015 & 0.009\,/\,0.002\\
         Python & 0.001 & 0.006\,/\,0.021 & 0.500 & 0.067\,/\,0.071 & 0.721 & 0.394\,/\,0.356 & 0.017 & 0.007\,/\,0.005 \\
         PHP  & 0.000 & 0.006\,/\,0.007 & 0.248 & 0.072\,/\,0.048 &  0.537 & 0.400\,/\,0.262 & 0.009 & 0.009\,/\,0.001\\
         \hline
         R & 0.011 & 0.005\,/\,0.004 & \multicolumn{2}{c|}{(no fine-tuning data)} &  0.045 & 0.139\,/\,0.118 & 0.024 & 0.002\,/\,0.002\\
         \bottomrule
    \end{tabular}
    \end{adjustbox}
    \label{tab:baseline}
\end{table}

\begin{table}[h!]
    %\small
    \centering
    \caption{MRR improvement as difference to non-whitened embeddings and IsoScores (Code\,/\,Comment) of the whitened embeddings using the best epsilon }
    \begin{adjustbox}{width=1\textwidth}
    \begin{tabular}{l|rr|rr|rr|rr}
        \toprule
         & 
         \multicolumn{2}{c}{CodeBERT} &
         \multicolumn{2}{c}{FT CodeBERT} &
         \multicolumn{2}{c}{CodeT5+} &
         \multicolumn{2}{c}{Code LLama} \\
         \midrule
         & $\Delta$MRR & IsoScores & $\Delta$MRR & IsoScores &$\Delta$ MRR & IsoScores & $\Delta$MRR & IsoScores \\
         \midrule
         Ruby & +0.230 & 0.365\,/\,0.511 & +0.075 & 0.998\,/\,0.998 & +0.007 & 0.377\,/\,0.340 & +0.476 & 0.224\,/\,0.298 \\
         Javascript  & +0.142 & 0.348\,/\,0.551 & +0.049 & 0.992\,/\,0.994 & +0.003 & 0.394\,/\,0.367 & +0.369 & 0.299\,/\,0.428 \\
         Go & +0.250 & 0.673\,/\,0.863 & +0.042 & 0.998\,/\,0.998 &  0.000 & 0.317\,/\,0.291 & +0.465 & 0.257\,/\,0.433 \\
         Java  & +0.148 & 0.736\,/\,0.889 & +0.064 & 0.991\,/\,0.992 & +0.002 & 0.522\,/\,0.495 & +0.329 & 0.453\,/\,0.388 \\
         Python & +0.156 & 0.381\,/\,0.555 & +0.062 & 0.998\,/\,0.998 & 0.000 & 0.420\,/\,0.386 & +0.399 & 0.326\,/\,0.496 \\
         PHP  & +0.102 & 0.726\,/\,0.899 & +0.055 & 0.998\,/\,0.998 & +0.002 & 0.423\,/\,0.327 & +0.227 & 0.275\,/\,0.450 \\
          \hline
         R & +0.077 & 0.462\,/\,0.352 & \multicolumn{2}{c|}{(no fine-tuning data)} &  +0.035 & 0.706\,/\,0.641 & +0.337 & 0.247\,/\,0.248 \\
         \bottomrule
    \end{tabular}
    \end{adjustbox}
    \label{tab:gains}
\end{table}

\section{Conclusion}
\label{sec:conclusion}
Controlling isotropy through Soft-ZCA whitening offers a simple yet effective way to improve code search performance across different code LMs and programming languages. The consistent MRR improvements suggest that embedding space geometry plays a crucial role in semantic code search. By improving the isotropy of embeddings, this post-processing technique offers a practical solution for enhancing code search systems in production.
% ****************************************************************************
% BIBLIOGRAPHY AREA
% ****************************************************************************

\bibliographystyle{unsrt}
\bibliography{refs}
% ****************************************************************************
% END OF BIBLIOGRAPHY AREA
% ****************************************************************************
%

%\newpage
\section{Appendix}

\subsection{CodeT5+ Embedding Size}

For our experiments with CodeT5+ we use the Hugging Face implementation of the \texttt{"Salesforce/codet5p-110m-embedding"} model. This adds an extra layer after the encoder to down-project the hidden size from 768 to 256 before applying mean pooling. In very high-dimensional spaces, data points can spread sparsely, making uniform coverage (and thus isotropy) harder to achieve. To test this hypothesis we compare using the last hidden state of the encoder to the default implementation. Table~\ref{tab:codet5p_emb} showcases the difference between using the two different embedding dimensions. Notably, the model with a smaller embedding size achieves nearly double the MRR and IsoScores, confirming the hypothesis that a lower dimensional embedding space has higher isotropy and translates to better downstream task performance.

\begin{table}[h!]
    \small
    \centering
    \caption{MRR and IsoScores (Code\,/\,Comment) of the non-whitened CodeT5+ embeddings with different hidden dimension size}
    \begin{tabular}{l|cc|cc}
        \toprule
         & 
         \multicolumn{2}{c}{Embedding Size 256} &
         \multicolumn{2}{c}{Embedding Size 768} \\
         \midrule
         & MRR & IsoScores & MRR & IsoScores\\
         \midrule
         Ruby &  0.705 & 0.350\,/\,0.296 & 0.463 & 0.102\,/\,0.102 \\
         Javascript   &  0.638 & 0.365\,/\,0.335 & 0.360 & 0.110\,/\,0.118 \\
         Go & 0.757 & 0.234\,/\,0.196& 0.413 & 0.042\,/\,0.044\\
         Java  & 0.595 & 0.388\,/\,0.313 & 0.323 & 0.120\,/\,0.081\\
         Python & 0.721 & 0.394\,/\,0.356 & 0.387 & 0.122\,/\,0.139 \\
         PHP  & 0.537 & 0.400\,/\,0.262 & 0.266 & 0.107\,/\,0.065\\
         \hline
         R & 0.045 & 0.139\,/\,0.118 & 0.026 & 0.026\,/\,0.012\\
         \bottomrule
    \end{tabular}
    \label{tab:codet5p_emb}
\end{table}

\subsection{Evaluating Separate vs Combined Whitening}
In our main experiments, we apply whitening to code and comments separately, as this approach closely reflects the structure of real production systems, where code and natural language query representations are processed independently. However, to assess the validity of this design choice, we also conduct experiments using combined whitening, where the two modalities are concatenated together for the whitening matrix calculation. This allows us to evaluate whether separate whitening truly provides an advantage or if a unified whitening matrix could yield comparable results. Table~\ref{tab:combined} showcases the difference between combines and seperate whitening, using standard ZCA whitening ($\epsilon=0$) on the CodeBERT embeddings. In most scenarios combined whitening achieves slightly lower MRR scores, demonstrating that separate whitening is not just practical but also superior to combined whitening.

\begin{table}[H]
    \small
    \centering
    \caption{MRR scores of the CodeBERT model embeddings with separate and combined ZCA whitening}
    \begin{adjustbox}{width=1\textwidth}
    \begin{tabular}{l|c|c|c|c}
        \toprule
         & \multicolumn{2}{c}{CodeBERT} & \multicolumn{2}{c}{FT CodeBERT} \\
         & 
         Separate Whitening &
         Combined Whitening & 
         Separate Whitening &
         Combined Whitening \\
         \midrule
         Ruby & 0.0959 & 0.1160 & 0.6225 & 0.6117 \\
         Javascript  & 0.1113 & 0.0883 & 0.4750 & 0.4693 \\
         Go & 0.0020 & 0.0037 & 0.5676 & 0.5557\\
         Java  & 0.0008 & 0.0006 & 0.4553 & 0.4482\\
         Python & 0.0396 & 0.0239 & 0.5524 & 0.5455 \\
         PHP  & 0.0005 & 0.0005 & 0.2847 & 0.2784 \\
         \hline
         R & 0.0712 & 0.0414 & \multicolumn{2}{c}{no fine-tuning data} \\
         \bottomrule
    \end{tabular}
    \end{adjustbox}
    \label{tab:combined}
\end{table}

\end{document}